\documentclass[sigconf]{acmart}

\usepackage{tabularx}
\usepackage{makecell}
\AtBeginDocument{%
  \providecommand\BibTeX{{%
    Bib\TeX}}}

\setcopyright{acmcopyright}

\usepackage{pifont}
\newcommand{\cmark}{\ding{51}}%
\usepackage{multirow}

\copyrightyear{2023}
\acmYear{2023}
\setcopyright{acmlicensed}\acmConference[MMSports '23]{Proceedings of the 6th International Workshop on Multimedia Content Analysis in Sports}{October 29, 2023}{Ottawa, ON, Canada}
\acmBooktitle{Proceedings of the 6th International Workshop on Multimedia Content Analysis in Sports (MMSports '23), October 29, 2023, Ottawa, ON, Canada}
\acmPrice{15.00}
\acmDOI{10.1145/3606038.3616172}
\acmISBN{979-8-4007-0269-3/23/10}

\begin{document}

\title{
Multi-task Learning for Joint Re-identification, Team Affiliation, and Role Classification for Sports Visual Tracking
}


\author{Amir M. Mansourian}
\affiliation{%
  \institution{Sharif University of Technology}
  \streetaddress{1 Th{\o}rv{\"a}ld Circle}
  \city{Tehran}
  \country{Iran}}
\email{amir.mansurian@sharif.edu}
\authornote{Authors contributed equally to this work.}

\author{Vladimir Somers}
\affiliation{%
  \institution{UCLouvain \& EPFL \& Sportradar}
  \streetaddress{1 Th{\o}rv{\"a}ld Circle}
  \city{Louvain-la-Neuve}
  \country{Belgium}}
\email{vladimir.somers@uclouvain.be}
\authornotemark[1]
\orcid{0000-0001-5787-4276}

\author{Christophe De Vleeschouwer}
\affiliation{%
  \institution{UCLouvain}
  \streetaddress{Place du Levant}
  \city{Louvain-la-Neuve}
  \country{Belgium}}
\email{christophe.devleeschouwer
@uclouvain.be}

\author{Shohreh Kasaei}
\affiliation{%
  \institution{Sharif University of Technology}
  \streetaddress{1 Th{\o}rv{\"a}ld Circle}
  \city{Tehran}
  \country{Iran}}
\email{kasaei@sharif.edu}

\begin{abstract}
Effective tracking and re-identification of players is essential for analyzing soccer videos. But, it is a challenging task due to the non-linear motion of players, the similarity in appearance of players from the same team, and frequent occlusions.
Therefore, the ability to extract meaningful embeddings to represent players is crucial in developing an effective tracking and re-identification system. In this paper, a multi-purpose part-based person representation method, called PRTreID, is proposed that performs three tasks of role classification, team affiliation, and re-identification, simultaneously. In contrast to available literature, a single network is trained with multi-task supervision to solve all three tasks, jointly. The proposed joint method is computationally efficient due to the shared backbone. Also, the multi-task learning leads to richer and more discriminative representations, as demonstrated by both quantitative and qualitative results. To demonstrate the effectiveness of PRTreID, it is integrated with a state-of-the-art tracking method, using a part-based post-processing module to handle long-term tracking.
The proposed tracking method, outperforms all existing tracking methods on the challenging SoccerNet tracking dataset. 
\end{abstract}



\maketitle

\begin{figure}[!ht]
\includegraphics[scale=.8]{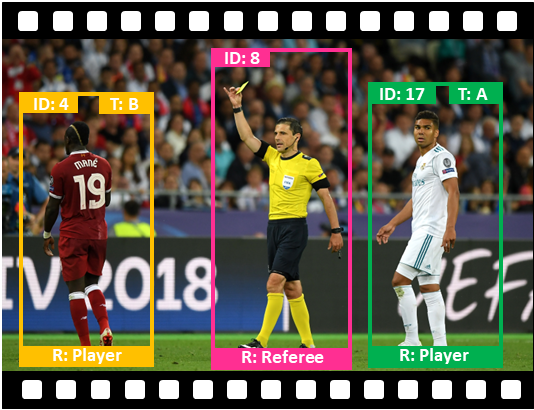}
\caption{Automated analytics for team sports requires tracking, re-identification, role classification (e.g. player, referee, staff, ...), and team affiliations (such as Team A or Team B) of all detected persons throughout an entire video of a game.}
\label{teaser}
\end{figure}
\section{Introduction}
 Automated sport analysis involves the use of computer vision and deep  learning techniques to analyze sports events and extract meaningful insights. It shows potential applications in a variety of areas, such as sports broadcasting \cite{vats2023player, tong2011automatic, cioppa2019arthus}, autonomous and personalized production \cite{chen2011formulating, Mkhallati2023SoccerNetCaptionDV, Cioppa2018ABA}, coaching \cite{chen2016learning, thomas2017computer, afrouzian2016pose, Held2023VARSVA}, and performance analysis \cite{Zandycke2022DeepSportradarv1CV, darwish2022ste, uchida2021automated, istasse2019associative, theagarajan2020automated, tavassolipour2013event}. One of the key components of automated sport analysis is object tracking, which involves the identification of players and objects all along a video. Tracking is essential for automated sport analysis for three main reasons. Firstly, it allows for the identification and distinction of players, enabling the extraction of personalized information. Secondly, it creates a spatio-temporal representation of the game, providing a detailed overview that can identify keys events, game patterns, and trends. Lastly, automatic tracking reduces the manual effort required for game strategy analysis, allowing analysts to focus on higher-level tasks. 

Tracking is challenging due to complex and dynamic motion, occlusions, inter-appearance similarities and intra-appearance variations. Re-identification, on the other hand, provides an important cue to solve tracking challenges after an occlusion or a move outside the camera field-of-view. 

Moreover, end-user applications typically demand that the upstream tracking system provides comprehensive high-level data regarding the individuals involved in sports, including roles (like player, referee, staff, goalkeeper, and so on) and team affiliations (such as Team A or B). Consequently, as illustrated in Figure \ref{teaser}, a desirable sports tracking system should simultaneously tackle tracking, re-identification, team affiliation, and role classification challanges.

Despite substantial advancements in sports analysis methods, as shown in recent studies \cite{Vandeghen2022SemiSupervisedTT, lu2013learning, hurault2020self, yang2021multi, maglo2022efficient, Ghasemzadeh2021DeepSportLabAU, Cioppa2022ScalingUS}, the majority of current tracking methods do not tackle all these tasks together. Solving each task individually is also not optimal as it overlooks the common objectives shared by all three tasks for accurately representing the individual, which could potentially benefit from a unified approach.
Most published sports tracking papers tend to concentrate on one or two tasks (like tracking and re-identification), often neglecting team and role identification, or the other way around.
Moreover, in those approaches, the role classification task is typically accomplished by an upstream object detector. Most state-of-the-art pre-trained detectors are limited to the "person" class and cannot distinguish between various sports roles (player, goalkeeper, referee, staff, etc). 
Similarly, the team affiliation task is often incorrectly formulated as a classification task with predefined teams \cite{vats2023player, zhang2020multi}, preventing the model to be used to distinguish new teams, unseen at the training phase. 
Finally, Re-Identification (ReID) models trained solely with identity labels may not be suitable for team affiliation, as they may consider players with similar attributes, such as skin color, more similar to each other than players from the same team, resulting in poor team clustering.

To address these limitations, a novel representation model is proposed that generates a multi-purpose representation from a single backbone, enabling to jointly solve the three tasks of person re-identification, team affiliation, and role classification.
The problems of re-identification and team affiliation are formulated as deep metric learning tasks, where players with the same identities are matched according to their similarity (i.e., distance) in the feature space. The team affiliation within a video is performed by clustering players' representations into two groups, thereby being applicable to the recognition of teams that were not seen at training. On the other hand, sports person's role prediction is formulated as a classification task.
Our method enjoys two key benefits; i) by integrating three distinct representation objectives during training, the model generates richer representations resulting in superior re-identification performance compared to models trained with a single objective and ii) it is efficient both in terms of speed and memory as it jointly solves three tasks using a single backbone and a single representation. 

In practice, a state-of-the-art body part-based re-identification (BPBreID) model \cite{somers2023body} is used as the baseline on top of which the proposed joint model is built by adding two objectives dedicated to role classification and team affiliation. 
A multi-task approach is then utilized to train the model jointly with all three objectives of re-identification, role classification, and team affiliation.
At inference, the model produces part-based features that can be used to solve these three tasks, simultaneously. The method is called \textit{PRTreID} for \textbf{P}art-based \textbf{R}ole classification \textbf{T}eam affiliation and person \textbf{re}-\textbf{ID}entification

Furthermore, the proposed PRTreID model is integrated with a state-of-the-art re-identification-based tracking method, namely StrongSORT \cite{du2023strongsort}, and the post-processing step of this tracker is replaced with the proposed part-based tracklet merging module.

In summary, the main contributions of this work are as follows:
\begin{itemize}
\item The PRTreID, a novel multi-task sports person representation model, is proposed to address re-identification, team affiliation, and role classification, simultaneously. 
To the best of our knowledge, this work is the first to address these three significant sport representation tasks with a single model and to demonstrate the benefit of multi-task learning in enhancing the richness of representation features.

\item The PRT-Track, a novel StrongSORT-based tracking method, is proposed to leverage the multi-task sports person representation model to produce long-term tracks.

\item Extensive experiments are conducted to validate the effectiveness of the proposed method and demonstrate the key benefits of multi-task learning. 

\end{itemize}

The prepared codebase and dataset will be released to encourage further research on joint representation learning for sports.

\begin{figure*}[!ht]
\centerline{\includegraphics[scale=.6]{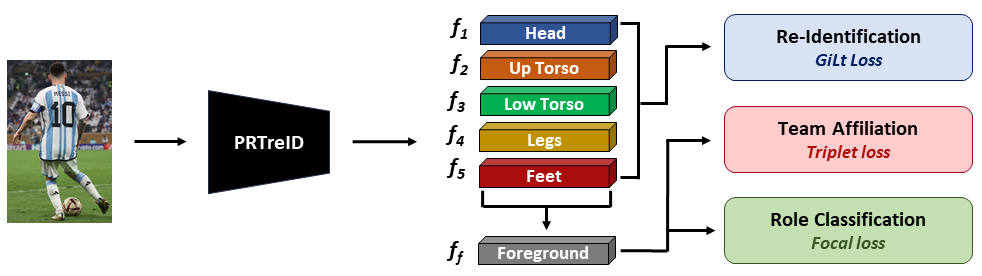}}
\caption{
Diagram of the proposed PRTreID method. An input image is fed into a shared backbone, which outputs an embedding for each part of the body. The foreground mask is created by combining all parts embeddings. The re-identification objective is trained with triplet loss and cross entropy loss on the body parts embeddings. Additionally, team affiliation and role classification objectives are trained with triplet loss and focal loss, respectively, on the foreground embedding. At inference time, the resulting multi-purpose embeddings can be utilized for person re-identification, team affiliation, and role classification tasks.}
\label{method_diagram}
\end{figure*}

\section{Related Work}
The literature most relevant to this work includes researches
surrounding the part-based re-identification and multiple object tracking.

\subsection{Part-Based Re-Identification}

Part-based re-identification methods have recently achieved state-of-the-art results in re-identification tasks, particularly in scenarios with occlusions, due to their ability to use local features for each part. However, one of the challenges faced by these methods is the need to localize each part accurately. Two different approaches are commonly used to address this problem. The first approach uses a fixed parts, which does not require any additional information for localizing parts \cite{sun2018beyond, xu2021dual, wang2021pose}. In the second approach, however, the method usually involves a pose estimation model to extract body parts, which is trained concurrently with the re-identification part \cite{suh2018part, kalayeh2018human}. For example, \cite{somers2023body} trains a model to find body parts of a person, using human parsing labels, and extracts embeddings for each part simultaneously. \cite{senocak2018part} proposes a part-based approach specific to sport works for player identification in basketball.

In the context of team sports, there are also other cues that can be helpful in re-identification of a person, such as team affiliation \cite{vats2023player, istasse2019associative, koshkina2021contrastive}, role information \cite{koshkina2021contrastive}, and jersey number \cite{vats2021multi, vats2022ice, liu2019pose}. \cite{istasse2019associative} trains a network that can output embeddings that are close for players on the same team and far from each other for players from different teams. In \cite{vats2023player} an automated system containing tracking, team classification, and jersey number recognition but used distinct networks to extracts those multiple information. In contrast, similar to our work, \cite{zhang2020multi} presents a deep player model for re-identification using jersey number, team class, and pose estimation features with a shared backbone. Our work goes one step further by adopting part-based representation features (and simply pose parameters), and by making those embedded features compliant with role and team affiliation prediction.

This paper utilizes the re-identification model introduced in \cite{somers2023body} to extract body-part-based features for persons in soccer videos, and adds two objectives for training the model: team affiliation and role classification. Training the model jointly for these three tasks leads to more meaningful part-based embeddings.

\subsection{Multiple Object Tracking}
The dominant paradigm in existing multiple object tracking methods is track-by-detection, where objects are detected in each frame before being associated across frames. SORT \cite{bewley2016simple} used a Kalman filter motion model and Intersection over Union (IoU) criteria for object association. Deep SORT \cite{wojke2017simple} improved upon SORT by adding appearance features alongside motion. Recent works such as OC-SORT \cite{cao2023observation} have made changes to SORT to handle non-linear motions, while others such as CBIoU \cite{yang2023hard} use a buffered version of IoU to extend bounding boxes for better object matching. More recently, re-identification-based trackers, like StrongSORT \cite{du2023strongsort}, have emerged, which focus on extracting more discriminative features for object re-identification to improve tracking results \cite{zhang2021fairmot, aharon2022bot, maggiolino2023deep}. Some works have also been specifically developed for tracking team sport players \cite{najafzadeh2015multiple, kc2016discriminative, maglo2022efficient, scott2022soccertrack, he2022application}. \cite{zhang2020multi} proposes a deep learning-based approach for multi-camera multi-player tracking in sports videos, leveraging deep player identification to improve tracking accuracy and consistency across multiple cameras. \cite{kc2016discriminative} addresses the specificity of use cases where appearance features are quite discriminant but only available sporadically, like it happens for the digits printed on the shirt of team sport players.

In this work, the Strong-SORT method \cite{du2023strongsort} is adopted. To improve the re-identification part, its appearance embeddings are replaced with part-based ones. Additionally, a post-processing step is proposed to further improve the tracking results by merging short tracklets with similar part-based features. 

\section{Proposed Methods}
In this section, first the body part-based re-identification baseline, initially proposed in \cite{somers2023body} is explained. Subsequently, two new components are introduced. They correspond to the \textit{team affiliation} training objective and to the multi-role classifier head. Finally, a tracking paradigm exploiting our proposed part-based features.

\subsection{PRTreID}
The overall architecture of our part-based ReID method \textit{PRTreID} is illustrated in Figure \ref{method_diagram}.
Compared to global Re-Identification (ReID) methods that produce a single embedding for each input sample, part-based methods produce multiple embeddings, each corresponding to a specific body region of the target person. Part-based methods have shown superior performance when facing occlusions \cite{PCB, PVPM} and are therefore a promising solution to address the multi-object tracking process. 
\subsubsection{\textbf{Part-based Re-Identification Baseline}}
In this work, the body part-based re-identification method, introduced in  \cite{somers2023body}, is employed as a baseline to re-identify persons in soccer videos. The so-called BPBReID model consists of a CNN feature extractor backbone, along with two modules for body-part attention and ReID, used for person re-identification. The backbone receives an input image and produces a spatial feature map ($G \in \mathbb{R}^{C \times H \times W}$) that is utilized in both modules. In the body part attention module, for each pixel $(h, w)$, a pixel-wise part classifier predicts whether this pixel belongs to the background or one of the $K$ body parts. This process is similar to a segmentation task for each body part, with $K+1$ classes corresponding to one of the $K$ body parts or to the background. This module employs a pixel-wise cross-entropy loss and utilizes human parsing labels (obtained from the Open Part Intensity Field, Part Association Field (OpenPifPaf) \cite{kreiss2021openpifpaf} pose estimation model) for training.
The overall loss to train this attention mechanism is called the \textit{part-prediction loss} $\ell_{pa}$, and is the sum of all pixel-wise cross-entropy loss on the spatial feature map G:

\begin{equation}
    \centering
\ell_{pa} = \sum_{h=0}^{H-1}\sum_{w=0}^{W-1}\ell_{CE}(G(h, w))
\end{equation}

This body part attention module produces two outputs for each body part: a mask ($M_k \in \mathbb{R}^{H\times W}; k \in \{1, ..., K\}$) and a binary visibility score ($v_k; k\in\{1, ..., K\}$) that indicates if the part is visible in the image. The masks of all parts are then merged to create a foreground mask, denoted $M_f$. The second module takes the outputs of the first module and performs a weighted average pooling of the spatial feature map using each attention mask, to produce $K+1$ embeddings;  i.e., one for each of the $K$ parts and the foreground ($f_k; k\in\{f, 1, ...,K\}$). In addition to these part-based embeddings, two other global features are extracted; one by concatenating all embeddings of $K$ parts ($f_c = concat(f_1, ...,f_K)$), and the other by performing the global average pooling on $G$ ($f_g$). 
Finally, the Global identity Local triplet (GiLt) loss, introduced in \cite{somers2023body}, is employed as the re-identification objective to train our model. GiLt is a re-identification loss designed especially for part-based ReID methods, that is robust to occluded and non-discriminative body parts:

\begin{equation} \label{bpb loss} 
    \centering
\ell_{reid} = \ell_{GiLt}(f_g, f_c, f_f, f_1, ..., f_K)
\end{equation}





The overall training objective of BPBReID is therefore the sum of two losses: the GiLt loss and the part prediction loss. 

Finally, at inference, the distance between two person's images "$q$" and "$g$" is computed as the average distance of their mutually visible body parts as 

{
\begin{equation} \label{eq:test_dist}
dist^{qg} = 
\frac{  \sum\limits_{i \in \{\text{f}, 1, ..., K\}} \Big(v^{q}_{i} \cdot v^{g}_{i} \cdot dist_{eucl}(f_{i}^{q}, f_{i}^{g}) \Big) }
{  \sum\limits_{i \in \{\text{f}, 1, ..., K\}} \big(v^{q}_{i} \cdot v^{g}_{i} \big) }\ ,
\end{equation}
}

where $v^{q|g}_{i}$ is the visibility score of body part $i$ and $dist_{eucl}$ is the Euclidean distance between two features.
A low distance $dist^{qg}$ between two person's images corresponds to a high similarity.

For more information about the BPBReID baseline, refer to \cite{somers2023body}.
In this work, the the aforementioned model is utilized to improve learning performance and extract richer player embeddings through the addition of a new role classification head and two new training objectives.


\subsubsection{\textbf{Role Classification}}
In the proposed method, a role classification head is added to the Re-Identification model. This new head is a fully connected layer for multi-class classification purposes. It is designed to classify individuals in soccer videos into four classes: player, goalkeeper, referee, and staff. The role classification head contributes to better representations and adds new semantics to the embeddings. The Focal loss ($\ell_{focal}$) \cite{lin2017focal} is employed to train it; because of the imbalance in the data between the player class and other classes. The role classification loss is denoted as $\ell_{role}$ and is applied on the foreground embedding.

\begin{equation} \label{role classification loss} 
    \centering
\ell_{role} = \ell_{focal}(f_f),
\end{equation}

\subsubsection{\textbf{Team Affiliation}}
An additional loss is added to the Re-Identification model. It aims at clustering players' embeddings according to their team affiliation. This loss function brings the embeddings of players from a same team closer to each other and pulls away the ones from distinct. This feature organization strategy is complementary to the ReID objective, which pulls together the embeddings from the same person and pulls away the embeddings of different persons. The triplet loss \cite{hoffer2015deep} is employed as the team affiliation loss. It is defined as

\begin{equation} \label{team affiliation loss} 
    \centering
\ell_{team} = \ell_{tri}(f_f),
\end{equation}
where players from the same team are assigned the same team ID and are considered as positive samples, while players from different teams are considered as negative samples. 
Last, but not least, this team affiliation loss is only applied to i) foreground embedding and ii) training samples having the "player" role; therefore, excluding goalkeepers, referees, and staff.

The proposed mehod enjoys the advantage of not restricting it to some predefined classes/teams, and therefore can be applied to any match with unseen teams. At the inference phase, a clustering algorithm with two clusters is conducted on the player's embeddings to assign their team label 

\subsubsection{\textbf{Overall Training Procedure}}
The final loss of the model is a combination of all the previously mentioned losses, and is defined as 
\begin{equation} \label{total loss} 
    \centering
\ell_{total} = \lambda_{pa}\ell_{pa} + \lambda_{reid}\ell_{reid} + \lambda_{team}\ell_{team} + \lambda_{role}\ell_{role},
\end{equation}
where $\ell_{total}$ is the total loss of the re-identification network, and $\lambda_{pa}$, $\lambda_{reid}$, $\lambda_{team}$, and $\lambda_{role}$ are hyperparameters that specify the scaling factors for each loss.

\subsection{PRT-Track}
In this section, the utilization of part-based player representation in supporting player tracking is described. The tracking method used as a baseline is first detailed, followed by an explanation of how both its online and offline modules are adapted to leverage the new part-based re-identification model.

\subsubsection{\textbf{StrongSORT Baseline}}
For multi-person tracking in soccer videos, the Strong-SORT \cite{du2023strongsort}, a recent method that uses an extended version of the Kalman filter to predict bounding boxes in the next frame, and a re-identification model to extract appearance features is employed. It generates a cost matrix using the IoU and the appearance similarity between the Kalman filter predictions and current detections. Linear assignment is then used in an online fashion to associate detections from the new frame with previous tracklets. The method also includes two lightweight post-processing modules; Appearance-Free Link (AFLink) and Gaussian-Smoothed Interpolation (GSI). In fact, AFLink performs global association with spatio-temporal information of tracklets instead of their appearance, while GSI is a Gaussian-smoothed interpolation that relieves missing detections.

\subsubsection{\textbf{Online Tracking Module}}
Our proposed PRTreID model is integrated into Strong-SORT by replacing its global ReID features with our part-based features. 
The detection-to-detection, detection-to-tracklet, and tracklet-to-tracklet ReID distances are computed using Eq.(\ref{eq:test_dist}). An exponential moving average (EMA) is applied to update the part-based features of the tracklet as

\begin{equation} \label{part-based EMA} 
    \centering
e_{k}^{t} = \alpha e_{k}^{t-1} \cdot v_{k}^{t-1} + ( 1 - \alpha) f_{k}^{t} \cdot v_{k}^{t} , \forall k \in \{f, 1, ..., K\},
\end{equation}

where $e_k^t$ and $f_k^t$ are the appearance features of the $k$-th body part of the tracklet and the matched detection, respectively. Parameter $\alpha$ is the momentum term and $v_k^{t}$ is the visibility score of the $k$-th part of the body, where $k \in \{f, 1, ..., K\}$.
The visibility score $v_k^{t-1}$ of body part $k$ of a tracklet at time $t-1$ is set to $1$ if the corresponding body part is visible in at least one of the detections that are part of the tracklet.

\subsubsection{\textbf{Offline Post-Processing Module}}
The output of the first online tracking part consists of a set of short tracklets that need further processings to be merged into long tracks. To address this problem, an additional \textit{offline} post-processing step based on part-based features is proposed to merge tracklets based on their underlying appearance in order to form long trajectories. To this end, a part-based appearance cost matrix for tracklets is build as 


\begin{equation} \label{track_merge}
A_{ij} = \begin{cases}
+\infty & i=j \\
dist_{total}^{ij} & \text{otherwise},
\end{cases}
\end{equation}

where $dist^{ij}$, the distance between tracklet $i$ and $j$, was introduced in Eq.(\ref{eq:test_dist}).
The matrix $A \in \mathbb{R}^{M\times M}$ is the appearance cost matrix and $M$ is the number of tracklets obtained from the first online tracking step. The linear assignment problem minimizing the sum of matching costs is then solved by the Hungarian algorithm. This post-processing step aims at improving the results and lower the identity switches. The output of this step is a set of complete trajectories, where each trajectory corresponds to a unique identity that has been tracked throughout the video sequence. 
The proposed part-based tracking method is called \textit{PRT-Track}.

Finally, it is worth mentioning that the team and role information are not explicitly used in the association stage. However, these details are implicitly encoded in the re-identification embeddings, owing to the multi-task training of the model. As a result, individuals belonging to different teams or having different roles are well-separated in the embedding space, with distances greater than the assigned association threshold, to prevent wrong associations.

\begin{table}[!htbp]
\caption{Some numbers for the proposed SoccerNet tracking-based re-identification dataset.}
\begin{center}
\begin{tabular}{ |c|c|c|c| } 
\hline
subset & \# ids & \# images & \# cameras \\
\hline
train & 1352 & 16217 & 57\\ 
query & 1146 & 3432 & 49\\ 
gallery & 1146 & 13719 & 49\\ 
\hline
\end{tabular}
\label{reid_dataset}
\end{center}
\end{table}

\begin{table}[!htbp]
  \small
  \centering
  \caption{Comparison of the proposed PRTreID method with some existing re-identification methods on the test set of our proposed re-identification dataset. R1 denotes rank-1 accuracy. Only PRTreID is designed to address all three tasks. PRTreID$_{team}$/PRTreID$_{role}$ refers to a variant of PRTreID where only the team/role objective is used at training.}
    \begin{tabular}{|c|c c|c c|c c|}
    \hline
   \multicolumn{1}{|c|}{\multirow{2}{*}{\makecell[c]{Method}}} & \multicolumn{2}{c}{Re-ID} & \multicolumn{2}{c}{Team Aff} & \multicolumn{2}{c|}{Role Cls} \\
    \cline{2-7}       & mAP & R1 & mAP & R1 & Acc & Prec \\
    \hline
    BoT \cite{luo2019bag} & 62.63 & \multicolumn{1}{c|}{80.24} & - & \multicolumn{1}{c|}{-} & - & - \\

    PCB \cite{sun2018beyond} & 63.61 & \multicolumn{1}{c|}{80.39} & - & \multicolumn{1}{c|}{-} & - & - \\

    BPBreID \cite{somers2023body} & 71.43 & \multicolumn{1}{c|}{89.31} & - & \multicolumn{1}{c|}{-} & - & - \\

     PRTreID$_{team}$ & - & \multicolumn{1}{c|}{-} & 91.48 & \multicolumn{1}{c|}{96.68} & - & - \\

     PRTreID$_{role}$ & - & \multicolumn{1}{c|}{-} & - & \multicolumn{1}{c|}{-} & 93.64 & 71.45 \\

    PRTreID & \textbf{72.59} & \multicolumn{1}{c|}{\textbf{89.57}} & \textbf{92.89} & \multicolumn{1}{c|}{\textbf{97.60}} & \textbf{94.27} & \textbf{74.36} \\

    \hline
    \end{tabular}%
  \label{reid_results}%
\end{table}%
\begin{table*}[!htbp]
\caption{Comparison of the overall tracking performance of the proposed part-based tracking method (PRT-Track) with recent tracking methods, from the 2022 SoccerNet Tracking challenge, on the test set of the SoccerNet-Tracking dataset using oracle detections and detections from YOLOv8. (The symbol $\dagger$ indicates that the results are reported from \cite{cioppa2022soccernet}).}
\begin{center}
\begin{tabular}{ |c|c|c|c|c|c|c|c| } 
\hline
\multicolumn{8}{|c|}{Oracle detections using ground truth} \\
\hline
Method & Setup & HOTA $\uparrow$ & DetA $\uparrow$ &  AssA $\uparrow$ & MOTA $\uparrow$ &  IDF1 $\uparrow$ &  IDs $\downarrow$ \\
\hline
DeepSORT$^{\dagger}$\cite{wojke2017simple} &  w/ GT & 69.52 & 82.62 & 58.66 & 94.84 & - & - \\ 

ByteTrack$^{\dagger}$\cite{zhang2022bytetrack} & w/ GT & 71.5 & 84.34 & 60.71 & 94.57 & -  & - \\ 

OC-SORT\cite{cao2023observation} & w/ GT & 80.94 & 97.81 & 66.98 & 96.76 & 74.79 & 6079  \\ 

StrongSORT\cite{du2023strongsort} & w/ GT &  83.75 & 95.08 & 73.78 & 94.67 & 79.13 & 2815 \\ 

StrongSORT++ & w/ GT &  84.08 & 95.07 & 74.36 & 94.62 & 79.76 & \textbf{2619} \\ 

CBIoU\cite{yang2023hard} & w/ GT &  89.20 & 99.40 & 80.00 & 99.40 & 86.10 & -\\ 

\textbf{PRT-Track} &  w/ GT & \textbf{90.77} & \textbf{99.83} & \textbf{82.53} & \textbf{98.66} & \textbf{88.47} & 3355\\ 

\hline
\hline
\multicolumn{8}{|c|}{YOLOv8 detections} \\
\hline
Method & Setup & HOTA $\uparrow$ & DetA $\uparrow$ &  AssA $\uparrow$ & MOTA $\uparrow$ &  IDF1 $\uparrow$ &  IDs $\downarrow$ \\
\hline
DeepSORT$^{\dagger}$\cite{wojke2017simple} & w/o GT & 36.63 & 40.02 & 33.76 & 33.91 & -  & -\\ 

FairMOT$^{\dagger}$ \cite{zhang2021fairmot}& w/o GT & 43.91 & 46.31 & 41.77 & 50.69 & - & -\\ 

ByteTrack$^{\dagger}$\cite{zhang2022bytetrack} & w/o GT &  47.22 & 44.49 & 50.25 & 31.74 & -  & -\\ 

OC-SORT\cite{cao2023observation} & w/o GT &  54.60 & \textbf{63.47} & 47.07 & \textbf{76.18} & 62.52  & 3593\\ 

StrongSORT\cite{du2023strongsort} & w/o GT &  54.86 & 62.19 & 48.79 & 74.52 & 65.1 & 2178\\ 

StrongSORT++ & w/o GT &  56.21 & 62.89 & 50.27 & 75.02 & 66.53 & 2106\\ 

\textbf{PRT-Track} & w/o GT &  \textbf{59.77} & 61.09 & \textbf{58.55} & 73.07 & \textbf{74.44} & \textbf{1428} \\ 
\hline

\end{tabular}
\label{tracking_results}
\end{center}
\end{table*}

\section{Experiments}
\begin{figure}[h]
\centerline{\includegraphics[scale=.35]{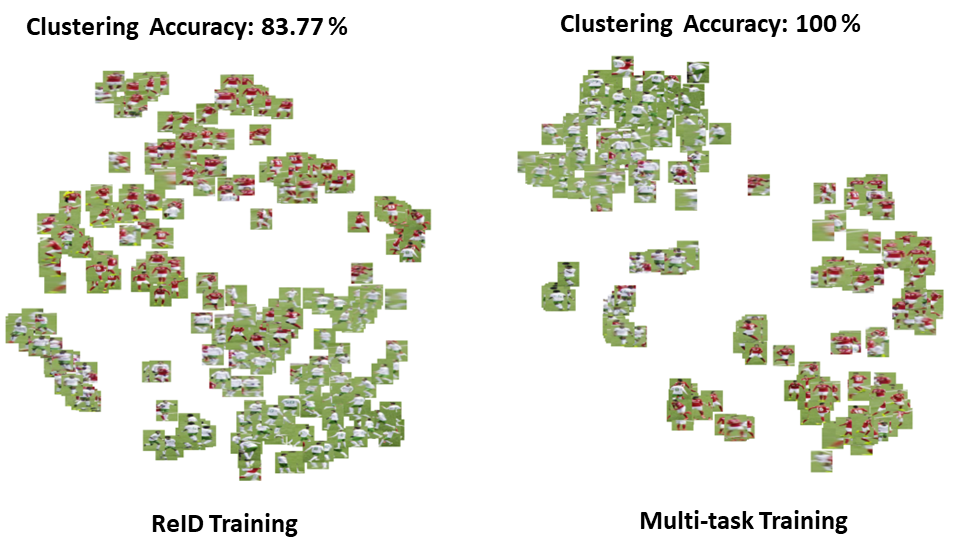}}
\caption{
t\_SNE visualization of player embeddings in a 2D space for a specific video, with and without multi-task training of the model. It can be observed that proposed multi-task model improves the team players clustering.}
\label{team_ablation_figure}
\end{figure}
\begin{table}[!ht]
\caption{Comparison of the post-processing techniques used by StrongSORT++ and our TrackMerge module (w/wo part-based features). }
\begin{center}

\begin{tabular}{|c|c|c|c|}
\hline
Post-processing method & HOTA & Det A & Ass A \\
\hline
StrongSORT++ & 84.08 & 95.07 & 74.36\\
\hline
TrackMerge + global features & 85.7 & 95.82 & 77.25 \\
\hline
TrackMerge + part-based features & 90.84 & 99.82 & 82.03\\
\hline
\end{tabular}

\label{tracking_ablation}
\end{center}
\end{table}

    


    
    

\begin{table*}[!ht]
  \centering
  \caption{Ablation for proposed PRTreID method to validate the benefits from multi-task learning.}
    \begin{tabular}{|c|c c c|c c|c c|c c|}
    \hline
    \multicolumn{1}{|c|}{\multirow{2}{*}{\makecell[c]{\#}}} & \multicolumn{3}{c|}{Loss} & \multicolumn{2}{c|}{Re-Identification} & \multicolumn{2}{c|}{Team Affiliation} & \multicolumn{2}{c|}{Role Classification} \\
    \cline{2-10}  
    & ReID & Team & Role &  mAP $\uparrow$ & Rank-1 $\uparrow$ & mAP $\uparrow$ & Rank-1 $\uparrow$ & Accuracy $\uparrow$ & Precision $\uparrow$ \\
    \hline
    1 & \cmark &   &   & 71.43 & \multicolumn{1}{c|}{89.31} & 87.12 & \multicolumn{1}{c|}{97.43} & - & - \\
    
    2 &   & \cmark &   & 12.34 & \multicolumn{1}{c|}{4.28} & 91.48 & \multicolumn{1}{c|}{96.68} & - & - \\

    3 &   &   & \cmark & 10.29 & \multicolumn{1}{c|}{4.28} & 55.38 & \multicolumn{1}{c|}{50.96} & 93.64 & 71.45 \\

    4 & \cmark & \cmark &   & 72.53  & \multicolumn{1}{c|}{89.04} & 89.03  & \multicolumn{1}{c|}{97.40} & - & -\\
    
    5 & \cmark &   & \cmark & \textbf{72.78}  & \multicolumn{1}{c|}{89.34} & 78.51  & \multicolumn{1}{c|}{97.33} & 93.27 & 68.79  \\
    
    6 & \cmark & \cmark & \cmark & 72.59  & \multicolumn{1}{c|}{\textbf{89.57}} & \textbf{92.89}  & \multicolumn{1}{c|}{\textbf{97.60}} & \textbf{94.27}   & \textbf{74.36} \\
    \hline
    \end{tabular}%
  \label{reid_ablation}%
\end{table*}%
\begin{figure}[h]
\centerline{\includegraphics[scale=.45]{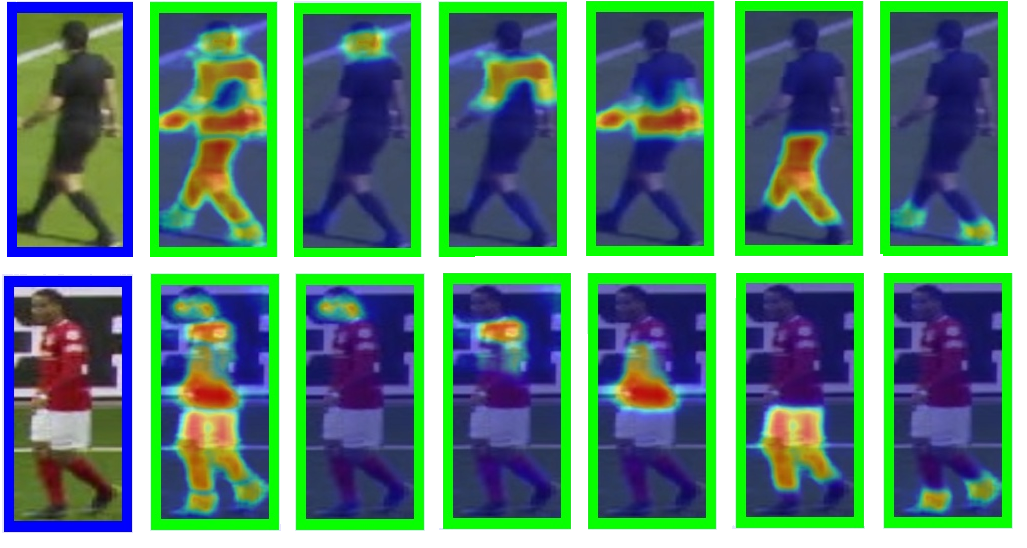}}
\caption{
Visualizations of two images from proposed ReID dataset and their attention maps of the foreground and each body part.}
\label{reid-vis}
\end{figure}

In this section, first, some details on the datasets, evaluation metrics, and implementation setups are given. Next, the detailed results of the proposed re-identification, team affiliation, role classification, and tracking methods are presented and analyzed. Finally, the ablation studies and qualitative assessments are discussed in order to further validate the effectiveness of the proposed method.

\begin{table}[!htbp]
\caption{Ablation study for team affiliation by applying a two-cluster Kmeans algorithm on foreground embedding of players for each video.}
\begin{center}
\begin{tabular}{  c | c  c c}
    \hline
     & ReID & ReID + Team & ReID + Team + Role   \\ \hline
    
     Accuracy & 91.87\% & 95.17\% \textcolor{red}{(+3.3)} & 95.6\% \textcolor{red}{(+3.73)} \\
    
    \hline
    \end{tabular}
\label{team_ablation_table}
\end{center}
\end{table}

\subsection{Datasets}

\subsubsection{SoccerNet Tracking}
SoccerNet-Tracking \cite{cioppa2022soccernet} is a publicly available dataset that contains player-tracking data from soccer matches. The dataset consists of 100 video clips, each 30 seconds long (25 frame per second), captured from the main camera. The dataset includes 57 videos from 3 games for the training set and 49 videos from 3 games for the test set. The objects in the dataset belong to the following classes: player team left, player team right, goalkeeper team left, goalkeeper team right, main referee, side referee, staff, and ball.

\subsubsection{Re-Identification Dataset} is used to train and evaluate the PRTreID model. A re-identification dataset was created by cropping objects (excluding the ball) from videos in the SoccerNet-Tracking dataset, using the same train/test splits. This resulted in a large number of detections per identity, much more than a standard re-identification dataset, with many of these detections being redundant and providing no additional information as they are from consecutive frames. To reduce the number of detections per identity, a uniform sampling approach was used along the frames. Table \ref{reid_dataset} presents the characteristics of the ReID dataset generated from SoccerNet-Tracking.

Finally, it is worth mentioning that the SoccerNet ReID \cite{giancola2022soccernet} dataset was not suitable for our research on video analysis and tracking, since it only includes cross-camera images of the same action without videos or cross-action annotations. This limits the ability to conduct large-scale experiments on team affiliation with multiple crops from two teams within a game video.

\subsection{Evaluation Metrics}
The performance of a given person re-identification method is typically evaluated as a retrieval problem: given an image of an individual of interest (the "\textit{query}") and a database of image crops of various persons (the "\textit{gallery}"), all gallery samples are ranked according to their distance to the query, such that gallery samples with the same identity as the query are at the top of the ranking. In line with standard evaluation practices in ReID research, the cumulative matching characteristics (CMC) at Rank-1 and the mean average precision (mAP) are used to assess the quality of query-gallery rankings. The same two metrics are also employed to evaluate team affiliation, using the same definition as in re-identification, reflecting the model's accuracy in retrieving images from the gallery set with the same \textit{team ID} as the query sample. For the role classification head, accuracy and precision are used as evaluation metrics.
In the tracking part, HOTA, IDF1, MOTA, AssA, ID Switches (IDs), and DetA are reported. Additionally, a two-class clustering is performed for the team affiliation part, and the model's accuracy in predicting team labels is reported.




\subsection{Implementation Details}
For the re-identification part, the BPBReID model was employed with the HRNet-W32 (HR) \cite{sun2019deep} backbone, which was pre-trained on the ImageNet dataset. The model was trained using weights from the Market 1501 \cite{zheng2015scalable} dataset and the number of parts (K) was set to 5 (head, up torso, low torso, legs, and feet) to achieve the best results. For the tracking part, the Strong-SORT tracker was utilized, and its feature extractor was replaced with the proposed PRTreID method.
The training configurations were the same as in a previous study \cite{somers2023body}. All images were resized to $384\times128$ and were first augmented with random cropping and 10-pixel padding, followed by random erasing at a 0.5 probability. All networks were trained end-to-end for 120 epochs using the Adam optimizer on a single NVIDIA GeForce RTX 3090 GPU. The learning rate was increased linearly from $3.5x10^{-5}$ to $3.5x10^{-4}$ after 10 epochs and was then decayed to $3.5x10^{-5}$ and $3.5x10^{-6}$ at the 40th and 70th epochs, respectively. A new batch sampler with a batch size of 32 was used for training. Each batch contained 4 identities from players of the left team, 4 identities from players of the right team, and 3 identities from other roles, with 4 images sampled for each identity which are all from a specific video. For the team affiliation objective, only identities from the player role were used, resulting in a batch size of 24. The triplet loss margin defined in Eq.(\ref{bpb loss}) was set to 0.3 and 0.05 for the re-identification and team affiliation objectives, respectively. The hyperparameters specified in Eq.(\ref{total loss}) were optimized by testing various values and selecting the optimal ones. Empirically, the values were set to 0.3, 1, 0.1, and 1.5 for $\lambda_{pa}$, $\lambda_{reid}$, $\lambda_{team}$, and $\lambda_{role}$ respectively.

\subsection{Experimental Results}

\subsubsection{Re-Identification Results}
Table \ref{reid_results} shows the performance of the proposed PRTreID model on the test set (query/gallery) of the proposed ReID dataset in comparison to some existing re-identification baselines. As this table shows, the proposed method outperforms other methods, despite the heavy occlusions in some images and the similar appearance of players from the same team. Additionally, PRTreID also performs well in terms of team affiliation and role classification. It should be noted that other methods do not support the other two tasks, which renders our model unique in its ability to perform multiple tasks simultaneously.
We could not compare the team affiliation and role classification performance of PRTreID with other works, because none provided open-source code allowing us to compute their performance on our dataset.

\subsubsection{Tracking Results}
In Table \ref{tracking_results},  the performance comparison of the proposed tracking method to existing tracking methods on the test set videos of the SoccerNet-Tracking dataset is given. Since our work focuses on feature learning and association for tracking, the results are compared to the trackers from the 2022 SoccerNet Tracking challenge. The proposed PRT-Track, with a part-based post-processing step, outperforms other methods on the SoccerNet-Tracking dataset, both with and without ground-truth detections. This demonstrates the importance of rich representations for tracking. Although it can be challenging to discriminate players by their appearance features in soccer due to the similar appearance of players and the distance between the main camera and the players, the proposed part-based tracking method achieves state-of-the-art performance and outperforms all previous methods.

\subsection{Ablation Studies}
To validate the effectiveness of each component of the PRTreID model, comprehensive ablation studies were conducted. Table \ref{reid_ablation} shows the impact of the two additional learning objectives on performance. As reported in the table, adding the team affiliation and role classification objectives during training improves the performance on both tasks as well as the re-identification performance. Additionally, we evaluate team affiliation performance at inference using a K-means clustering with two clusters on the foreground embedding of players for each video. After clustering, a team label was assigned to each player. The results in Table \ref{team_ablation_table} demonstrate the improvements of multi-task training on the overall model performance; the joint model outperforms all other partial models on all tasks. Finally, Table \ref{tracking_ablation} reports a comparison of the proposed appearance-based tracklet merging step, with those of StrongSORT++ (AFLink and GSI). The results show that the proposed tracklet merging with global features outperforms the StrongSORT++. Also, the tracklet merging with part-based features achieves the best results.

\begin{figure}[!ht]
\centerline{\includegraphics[scale=.4]{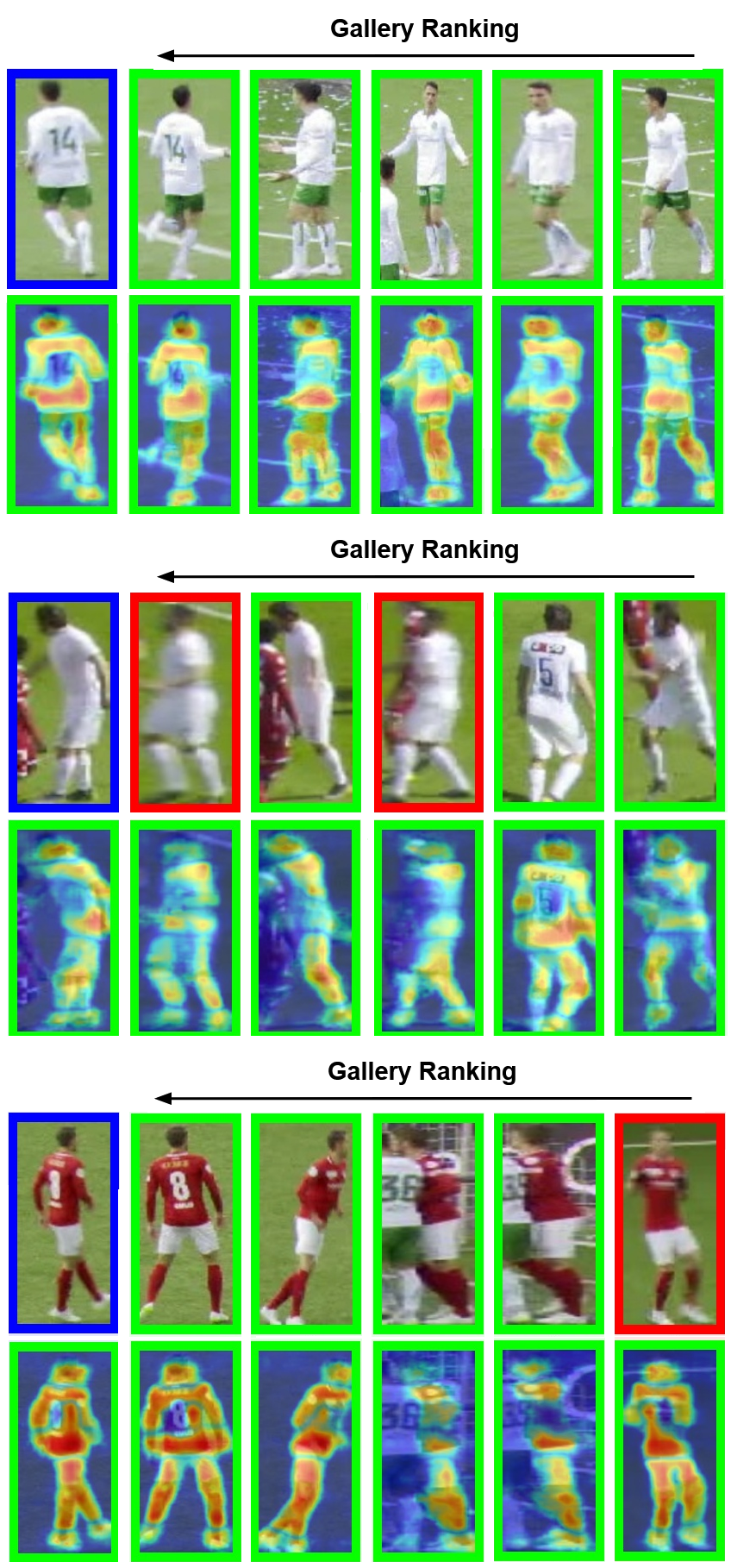}}
\caption{
Visualization of three images in the query set and their top-5 retrieved images from the gallery set of proposed ReID dataset, along with foreground masks. The blue color represents the query sample, while the red color indicates a wrongly retrieved image and the green color indicates a correctly retrieved image from the gallery.}
\label{ranking-vis}
\end{figure}

\subsection{Qualitative Assessment}
To further validate the effectiveness of the proposed method, some qualitative assessments of the model's performance are conducted. Figure \ref{reid-vis} illustrates two images from the proposed ReID dataset and their respective output attention maps. Each attention map highlights a different part; five parts of the body and the foreground. As illustrated here, the part attention module is still performing well when trained with other objectives; as it correctly extracts body parts heatmaps while excluding the background. Figure \ref{reid-vis} also shows three samples of query set and their corresponding top 5 retrieved gallery images from the proposed ReID dataset, along with their foreground attention maps. The red color indicates that the retrieved image has a different identity than the query image. This visualization demonstrates that in occluded scenarios, PRTreID can properly extract the attention maps of the target person and retrieve the correct images.

Figure \ref{team_ablation_figure} shows qualitative results of the learned embedding space, that is reduced to 2 dimensions using the t-SNE algorithm. As shown in the figure, the embeddings of different teams are much better distributed in the embedding space when the model is trained with all three objectives than with just the re-identification objective.
In summary, these qualitative results demonstrate that using multi-task learning to train the model leads to richer and more discriminative representations.

\subsection{Computational Efficiency Discussion}
For the readers that are concerned about the computational efficiency of our PRTreID model: our model has two main components: a backbone and a ReID head comprising an attention mechanism for part-based feature extraction and a classifier to perform role classification. Since the ReID head induces a very small computational overhead, the overall inference time of PRTreID will be almost the same as the backbone itself. In our work, we use a HRNet-W32 for its high-resolution feature maps that are beneficial to ReID, but any other smaller backbone can be used depending on the speed requirement of the downstream application.

\section{Conclusion and Future Work}
In this work, a novel person representation method was proposed for sports videos. A body part-based person re-identification model was utilized for extracting person embeddings. To extract more meaningful embeddings, two new objectives were added to the re-identification model for team affiliation and role classification purposes. Experimental results showed that jointly training the re-id model for re-identification, team affiliation, and role classification led to rich representations, which were then used in tracking persons as a downstream task of re-identification. Additionally, a post-processing step was proposed using part-based features to further improve the tracking results. The experimental results showed the superiority of the proposed method over the existing state-of-the-art methods.
In this study, our evaluation was limited to the SoccerNet tracking dataset, as we couldn't find any other open-source datasets providing annotations for re-identification, team affiliation, and role classification all at once. 
In future work, we plan to integrate a Jersey Number Recognition head into our sports person representation model. This decision is motivated by the challenges in team sports like ice hockey, where identical kits covering a large portion of the player's body can compromise player re-identification.


\begin{acks}
The authors would like to acknowledge the High Performance Center (HPC) of Sharif University of Technology for providing the computational resources for this project.
The authors would also like to thank Sportradar AG for sponsoring this work and for its commitment to the scientific research, making it possible for us to make significant contributions to the sports technology field.
\end{acks}
\bibliographystyle{ACM-Reference-Format}
\bibliography{main}


\begin{thebibliography}{58}


\ifx \showCODEN    \undefined \def \showCODEN     #1{\unskip}     \fi
\ifx \showDOI      \undefined \def \showDOI       #1{#1}\fi
\ifx \showISBNx    \undefined \def \showISBNx     #1{\unskip}     \fi
\ifx \showISBNxiii \undefined \def \showISBNxiii  #1{\unskip}     \fi
\ifx \showISSN     \undefined \def \showISSN      #1{\unskip}     \fi
\ifx \showLCCN     \undefined \def \showLCCN      #1{\unskip}     \fi
\ifx \shownote     \undefined \def \shownote      #1{#1}          \fi
\ifx \showarticletitle \undefined \def \showarticletitle #1{#1}   \fi
\ifx \showURL      \undefined \def \showURL       {\relax}        \fi
\providecommand\bibfield[2]{#2}
\providecommand\bibinfo[2]{#2}
\providecommand\natexlab[1]{#1}
\providecommand\showeprint[2][]{arXiv:#2}

\bibitem[Afrouzian et~al\mbox{.}(2016)]%
        {afrouzian2016pose}
\bibfield{author}{\bibinfo{person}{Reza Afrouzian}, \bibinfo{person}{Hadi
  Seyedarabi}, {and} \bibinfo{person}{Shohreh Kasaei}.}
  \bibinfo{year}{2016}\natexlab{}.
\newblock \showarticletitle{Pose estimation of soccer players using multiple
  uncalibrated cameras}.
\newblock \bibinfo{journal}{\emph{Multimedia Tools and Applications}}
  \bibinfo{volume}{75} (\bibinfo{year}{2016}), \bibinfo{pages}{6809--6827}.
\newblock


\bibitem[Aharon et~al\mbox{.}(2022)]%
        {aharon2022bot}
\bibfield{author}{\bibinfo{person}{Nir Aharon}, \bibinfo{person}{Roy Orfaig},
  {and} \bibinfo{person}{Ben-Zion Bobrovsky}.} \bibinfo{year}{2022}\natexlab{}.
\newblock \showarticletitle{BoT-SORT: Robust associations multi-pedestrian
  tracking}.
\newblock \bibinfo{journal}{\emph{arXiv preprint arXiv:2206.14651}}
  (\bibinfo{year}{2022}).
\newblock


\bibitem[Bewley et~al\mbox{.}(2016)]%
        {bewley2016simple}
\bibfield{author}{\bibinfo{person}{Alex Bewley}, \bibinfo{person}{Zongyuan Ge},
  \bibinfo{person}{Lionel Ott}, \bibinfo{person}{Fabio Ramos}, {and}
  \bibinfo{person}{Ben Upcroft}.} \bibinfo{year}{2016}\natexlab{}.
\newblock \showarticletitle{Simple online and realtime tracking}. In
  \bibinfo{booktitle}{\emph{2016 IEEE international conference on image
  processing (ICIP)}}. IEEE, \bibinfo{pages}{3464--3468}.
\newblock


\bibitem[Cao et~al\mbox{.}(2023)]%
        {cao2023observation}
\bibfield{author}{\bibinfo{person}{Jinkun Cao}, \bibinfo{person}{Jiangmiao
  Pang}, \bibinfo{person}{Xinshuo Weng}, \bibinfo{person}{Rawal Khirodkar},
  {and} \bibinfo{person}{Kris Kitani}.} \bibinfo{year}{2023}\natexlab{}.
\newblock \showarticletitle{Observation-centric sort: Rethinking sort for
  robust multi-object tracking}. In \bibinfo{booktitle}{\emph{Proceedings of
  the IEEE/CVF Conference on Computer Vision and Pattern Recognition}}.
  \bibinfo{pages}{9686--9696}.
\newblock


\bibitem[Chen and De~Vleeschouwer(2011)]%
        {chen2011formulating}
\bibfield{author}{\bibinfo{person}{Fan Chen} {and} \bibinfo{person}{Christophe
  De~Vleeschouwer}.} \bibinfo{year}{2011}\natexlab{}.
\newblock \showarticletitle{Formulating team-sport video summarization as a
  resource allocation problem}.
\newblock \bibinfo{journal}{\emph{IEEE Transactions on Circuits and Systems for
  Video Technology}} \bibinfo{volume}{21}, \bibinfo{number}{2}
  (\bibinfo{year}{2011}), \bibinfo{pages}{193--205}.
\newblock


\bibitem[Chen et~al\mbox{.}(2016)]%
        {chen2016learning}
\bibfield{author}{\bibinfo{person}{Jianhui Chen}, \bibinfo{person}{Hoang~M Le},
  \bibinfo{person}{Peter Carr}, \bibinfo{person}{Yisong Yue}, {and}
  \bibinfo{person}{James~J Little}.} \bibinfo{year}{2016}\natexlab{}.
\newblock \showarticletitle{Learning online smooth predictors for realtime
  camera planning using recurrent decision trees}. In
  \bibinfo{booktitle}{\emph{Proceedings of the IEEE Conference on Computer
  Vision and Pattern Recognition}}. \bibinfo{pages}{4688--4696}.
\newblock


\bibitem[Cioppa et~al\mbox{.}(2018)]%
        {Cioppa2018ABA}
\bibfield{author}{\bibinfo{person}{Anthony Cioppa}, \bibinfo{person}{Adrien
  Deli{\`e}ge}, {and} \bibinfo{person}{Marc~Van Droogenbroeck}.}
  \bibinfo{year}{2018}\natexlab{}.
\newblock \showarticletitle{A Bottom-Up Approach Based on Semantics for the
  Interpretation of the Main Camera Stream in Soccer Games}.
\newblock \bibinfo{journal}{\emph{2018 IEEE/CVF Conference on Computer Vision
  and Pattern Recognition Workshops (CVPRW)}} (\bibinfo{year}{2018}),
  \bibinfo{pages}{1846--184609}.
\newblock
\urldef\tempurl%
\url{https://api.semanticscholar.org/CorpusID:52027492}
\showURL{%
\tempurl}


\bibitem[Cioppa et~al\mbox{.}(2022a)]%
        {Cioppa2022ScalingUS}
\bibfield{author}{\bibinfo{person}{Anthony Cioppa}, \bibinfo{person}{Adrien
  Deli{\`e}ge}, \bibinfo{person}{Silvio Giancola}, \bibinfo{person}{Bernard
  Ghanem}, {and} \bibinfo{person}{Marc~Van Droogenbroeck}.}
  \bibinfo{year}{2022}\natexlab{a}.
\newblock \showarticletitle{Scaling up SoccerNet with multi-view spatial
  localization and re-identification}.
\newblock \bibinfo{journal}{\emph{Scientific Data}}  \bibinfo{volume}{9}
  (\bibinfo{year}{2022}).
\newblock
\urldef\tempurl%
\url{https://api.semanticscholar.org/CorpusID:249894209}
\showURL{%
\tempurl}


\bibitem[Cioppa et~al\mbox{.}(2019)]%
        {cioppa2019arthus}
\bibfield{author}{\bibinfo{person}{Anthony Cioppa}, \bibinfo{person}{Adrien
  Deliege}, \bibinfo{person}{Maxime Istasse}, \bibinfo{person}{Christophe
  De~Vleeschouwer}, {and} \bibinfo{person}{Marc Van~Droogenbroeck}.}
  \bibinfo{year}{2019}\natexlab{}.
\newblock \showarticletitle{ARTHuS: Adaptive real-time human segmentation in
  sports through online distillation}. In \bibinfo{booktitle}{\emph{Proceedings
  of the IEEE/CVF Conference on Computer Vision and Pattern Recognition
  Workshops}}. \bibinfo{pages}{0--0}.
\newblock


\bibitem[Cioppa et~al\mbox{.}(2022b)]%
        {cioppa2022soccernet}
\bibfield{author}{\bibinfo{person}{Anthony Cioppa}, \bibinfo{person}{Silvio
  Giancola}, \bibinfo{person}{Adrien Deliege}, \bibinfo{person}{Le Kang},
  \bibinfo{person}{Xin Zhou}, \bibinfo{person}{Zhiyu Cheng},
  \bibinfo{person}{Bernard Ghanem}, {and} \bibinfo{person}{Marc
  Van~Droogenbroeck}.} \bibinfo{year}{2022}\natexlab{b}.
\newblock \showarticletitle{SoccerNet-Tracking: Multiple Object Tracking
  Dataset and Benchmark in Soccer Videos}. In
  \bibinfo{booktitle}{\emph{Proceedings of the IEEE/CVF Conference on Computer
  Vision and Pattern Recognition}}. \bibinfo{pages}{3491--3502}.
\newblock


\bibitem[Darwish and El-Shabrway(2022)]%
        {darwish2022ste}
\bibfield{author}{\bibinfo{person}{Abdulrahman Darwish} {and}
  \bibinfo{person}{Tallal El-Shabrway}.} \bibinfo{year}{2022}\natexlab{}.
\newblock \showarticletitle{STE: Spatio-Temporal Encoder for Action Spotting in
  Soccer Videos}. In \bibinfo{booktitle}{\emph{Proceedings of the 5th
  International ACM Workshop on Multimedia Content Analysis in Sports}}.
  \bibinfo{pages}{87--92}.
\newblock


\bibitem[Du et~al\mbox{.}(2023)]%
        {du2023strongsort}
\bibfield{author}{\bibinfo{person}{Yunhao Du}, \bibinfo{person}{Zhicheng Zhao},
  \bibinfo{person}{Yang Song}, \bibinfo{person}{Yanyun Zhao},
  \bibinfo{person}{Fei Su}, \bibinfo{person}{Tao Gong}, {and}
  \bibinfo{person}{Hongying Meng}.} \bibinfo{year}{2023}\natexlab{}.
\newblock \showarticletitle{Strongsort: Make deepsort great again}.
\newblock \bibinfo{journal}{\emph{IEEE Transactions on Multimedia}}
  (\bibinfo{year}{2023}).
\newblock


\bibitem[Gao et~al\mbox{.}(2020)]%
        {PVPM}
\bibfield{author}{\bibinfo{person}{Shang Gao}, \bibinfo{person}{Jingya Wang},
  \bibinfo{person}{Huchuan Lu}, {and} \bibinfo{person}{Zimo Liu}.}
  \bibinfo{year}{2020}\natexlab{}.
\newblock \showarticletitle{{Pose-guided visible part matching for occluded
  person ReID}}. In \bibinfo{booktitle}{\emph{Proceedings of the IEEE Computer
  Society Conference on Computer Vision and Pattern Recognition}}.
  \bibinfo{pages}{11741--11749}.
\newblock
\showISSN{10636919}
\urldef\tempurl%
\url{https://doi.org/10.1109/CVPR42600.2020.01176}
\showDOI{\tempurl}
\showeprint[arxiv]{2004.00230}


\bibitem[Ghasemzadeh et~al\mbox{.}(2021)]%
        {Ghasemzadeh2021DeepSportLabAU}
\bibfield{author}{\bibinfo{person}{Seyed~Abolfazl Ghasemzadeh},
  \bibinfo{person}{Gabriel~Van Zandycke}, \bibinfo{person}{Maxime Istasse},
  \bibinfo{person}{Niels Sayez}, \bibinfo{person}{Amirafshar Moshtaghpour},
  {and} \bibinfo{person}{Christophe~De Vleeschouwer}.}
  \bibinfo{year}{2021}\natexlab{}.
\newblock \showarticletitle{DeepSportLab: a Unified Framework for Ball
  Detection, Player Instance Segmentation and Pose Estimation in Team Sports
  Scenes}.
\newblock \bibinfo{journal}{\emph{ArXiv}}  \bibinfo{volume}{abs/2112.00627}
  (\bibinfo{year}{2021}).
\newblock
\urldef\tempurl%
\url{https://api.semanticscholar.org/CorpusID:244773081}
\showURL{%
\tempurl}


\bibitem[Giancola et~al\mbox{.}(2022)]%
        {giancola2022soccernet}
\bibfield{author}{\bibinfo{person}{Silvio Giancola}, \bibinfo{person}{Anthony
  Cioppa}, \bibinfo{person}{Adrien Deli{\`e}ge}, \bibinfo{person}{Floriane
  Magera}, \bibinfo{person}{Vladimir Somers}, \bibinfo{person}{Le Kang},
  \bibinfo{person}{Xin Zhou}, \bibinfo{person}{Olivier Barnich},
  \bibinfo{person}{Christophe De~Vleeschouwer}, \bibinfo{person}{Alexandre
  Alahi}, {et~al\mbox{.}}} \bibinfo{year}{2022}\natexlab{}.
\newblock \showarticletitle{SoccerNet 2022 challenges results}. In
  \bibinfo{booktitle}{\emph{Proceedings of the 5th International ACM Workshop
  on Multimedia Content Analysis in Sports}}. \bibinfo{pages}{75--86}.
\newblock


\bibitem[He(2022)]%
        {he2022application}
\bibfield{author}{\bibinfo{person}{Xin He}.} \bibinfo{year}{2022}\natexlab{}.
\newblock \showarticletitle{Application of deep learning in video target
  tracking of soccer players}.
\newblock \bibinfo{journal}{\emph{Soft Computing}} \bibinfo{volume}{26},
  \bibinfo{number}{20} (\bibinfo{year}{2022}), \bibinfo{pages}{10971--10979}.
\newblock


\bibitem[Held et~al\mbox{.}(2023)]%
        {Held2023VARSVA}
\bibfield{author}{\bibinfo{person}{Jan Held}, \bibinfo{person}{Anthony Cioppa},
  \bibinfo{person}{Silvio Giancola}, \bibinfo{person}{Abdullah Hamdi},
  \bibinfo{person}{Bernard Ghanem}, {and} \bibinfo{person}{Marc~Van
  Droogenbroeck}.} \bibinfo{year}{2023}\natexlab{}.
\newblock \showarticletitle{VARS: Video Assistant Referee System for Automated
  Soccer Decision Making from Multiple Views}.
\newblock \bibinfo{journal}{\emph{2023 IEEE/CVF Conference on Computer Vision
  and Pattern Recognition Workshops (CVPRW)}} (\bibinfo{year}{2023}),
  \bibinfo{pages}{5086--5097}.
\newblock
\urldef\tempurl%
\url{https://api.semanticscholar.org/CorpusID:258048692}
\showURL{%
\tempurl}


\bibitem[Hoffer and Ailon(2015)]%
        {hoffer2015deep}
\bibfield{author}{\bibinfo{person}{Elad Hoffer} {and} \bibinfo{person}{Nir
  Ailon}.} \bibinfo{year}{2015}\natexlab{}.
\newblock \showarticletitle{Deep metric learning using triplet network}. In
  \bibinfo{booktitle}{\emph{Similarity-Based Pattern Recognition: Third
  International Workshop, SIMBAD 2015, Copenhagen, Denmark, October 12-14,
  2015. Proceedings 3}}. Springer, \bibinfo{pages}{84--92}.
\newblock


\bibitem[Hurault et~al\mbox{.}(2020)]%
        {hurault2020self}
\bibfield{author}{\bibinfo{person}{Samuel Hurault}, \bibinfo{person}{Coloma
  Ballester}, {and} \bibinfo{person}{Gloria Haro}.}
  \bibinfo{year}{2020}\natexlab{}.
\newblock \showarticletitle{Self-supervised small soccer player detection and
  tracking}. In \bibinfo{booktitle}{\emph{Proceedings of the 3rd international
  workshop on multimedia content analysis in sports}}. \bibinfo{pages}{9--18}.
\newblock


\bibitem[Istasse et~al\mbox{.}(2019)]%
        {istasse2019associative}
\bibfield{author}{\bibinfo{person}{Maxime Istasse}, \bibinfo{person}{Julien
  Moreau}, {and} \bibinfo{person}{Christophe De~Vleeschouwer}.}
  \bibinfo{year}{2019}\natexlab{}.
\newblock \showarticletitle{Associative embedding for team discrimination}. In
  \bibinfo{booktitle}{\emph{Proceedings of the IEEE/CVF Conference on Computer
  Vision and Pattern Recognition Workshops}}. \bibinfo{pages}{0--0}.
\newblock


\bibitem[Kalayeh et~al\mbox{.}(2018)]%
        {kalayeh2018human}
\bibfield{author}{\bibinfo{person}{Mahdi~M Kalayeh}, \bibinfo{person}{Emrah
  Basaran}, \bibinfo{person}{Muhittin G{\"o}kmen}, \bibinfo{person}{Mustafa~E
  Kamasak}, {and} \bibinfo{person}{Mubarak Shah}.}
  \bibinfo{year}{2018}\natexlab{}.
\newblock \showarticletitle{Human semantic parsing for person
  re-identification}. In \bibinfo{booktitle}{\emph{Proceedings of the IEEE
  conference on computer vision and pattern recognition}}.
  \bibinfo{pages}{1062--1071}.
\newblock


\bibitem[KC et~al\mbox{.}(2016)]%
        {kc2016discriminative}
\bibfield{author}{\bibinfo{person}{Amit~Kumar KC}, \bibinfo{person}{Laurent
  Jacques}, {and} \bibinfo{person}{Christophe De~Vleeschouwer}.}
  \bibinfo{year}{2016}\natexlab{}.
\newblock \showarticletitle{Discriminative and efficient label propagation on
  complementary graphs for multi-object tracking}.
\newblock \bibinfo{journal}{\emph{IEEE transactions on pattern analysis and
  machine intelligence}} \bibinfo{volume}{39}, \bibinfo{number}{1}
  (\bibinfo{year}{2016}), \bibinfo{pages}{61--74}.
\newblock


\bibitem[Koshkina et~al\mbox{.}(2021)]%
        {koshkina2021contrastive}
\bibfield{author}{\bibinfo{person}{Maria Koshkina}, \bibinfo{person}{Hemanth
  Pidaparthy}, {and} \bibinfo{person}{James~H Elder}.}
  \bibinfo{year}{2021}\natexlab{}.
\newblock \showarticletitle{Contrastive learning for sports video: Unsupervised
  player classification}. In \bibinfo{booktitle}{\emph{Proceedings of the
  IEEE/CVF Conference on Computer Vision and Pattern Recognition}}.
  \bibinfo{pages}{4528--4536}.
\newblock


\bibitem[Kreiss et~al\mbox{.}(2021)]%
        {kreiss2021openpifpaf}
\bibfield{author}{\bibinfo{person}{Sven Kreiss}, \bibinfo{person}{Lorenzo
  Bertoni}, {and} \bibinfo{person}{Alexandre Alahi}.}
  \bibinfo{year}{2021}\natexlab{}.
\newblock \showarticletitle{Openpifpaf: Composite fields for semantic keypoint
  detection and spatio-temporal association}.
\newblock \bibinfo{journal}{\emph{IEEE Transactions on Intelligent
  Transportation Systems}} \bibinfo{volume}{23}, \bibinfo{number}{8}
  (\bibinfo{year}{2021}), \bibinfo{pages}{13498--13511}.
\newblock


\bibitem[Lin et~al\mbox{.}(2017)]%
        {lin2017focal}
\bibfield{author}{\bibinfo{person}{Tsung-Yi Lin}, \bibinfo{person}{Priya
  Goyal}, \bibinfo{person}{Ross Girshick}, \bibinfo{person}{Kaiming He}, {and}
  \bibinfo{person}{Piotr Doll{\'a}r}.} \bibinfo{year}{2017}\natexlab{}.
\newblock \showarticletitle{Focal loss for dense object detection}. In
  \bibinfo{booktitle}{\emph{Proceedings of the IEEE international conference on
  computer vision}}. \bibinfo{pages}{2980--2988}.
\newblock


\bibitem[Liu and Bhanu(2019)]%
        {liu2019pose}
\bibfield{author}{\bibinfo{person}{Hengyue Liu} {and} \bibinfo{person}{Bir
  Bhanu}.} \bibinfo{year}{2019}\natexlab{}.
\newblock \showarticletitle{Pose-guided R-CNN for jersey number recognition in
  sports}. In \bibinfo{booktitle}{\emph{Proceedings of the IEEE/CVF Conference
  on Computer Vision and Pattern Recognition Workshops}}.
  \bibinfo{pages}{0--0}.
\newblock


\bibitem[Lu et~al\mbox{.}(2013)]%
        {lu2013learning}
\bibfield{author}{\bibinfo{person}{Wei-Lwun Lu}, \bibinfo{person}{Jo-Anne
  Ting}, \bibinfo{person}{James~J Little}, {and} \bibinfo{person}{Kevin~P
  Murphy}.} \bibinfo{year}{2013}\natexlab{}.
\newblock \showarticletitle{Learning to track and identify players from
  broadcast sports videos}.
\newblock \bibinfo{journal}{\emph{IEEE transactions on pattern analysis and
  machine intelligence}} \bibinfo{volume}{35}, \bibinfo{number}{7}
  (\bibinfo{year}{2013}), \bibinfo{pages}{1704--1716}.
\newblock


\bibitem[Luo et~al\mbox{.}(2019)]%
        {luo2019bag}
\bibfield{author}{\bibinfo{person}{Hao Luo}, \bibinfo{person}{Youzhi Gu},
  \bibinfo{person}{Xingyu Liao}, \bibinfo{person}{Shenqi Lai}, {and}
  \bibinfo{person}{Wei Jiang}.} \bibinfo{year}{2019}\natexlab{}.
\newblock \showarticletitle{Bag of tricks and a strong baseline for deep person
  re-identification}. In \bibinfo{booktitle}{\emph{Proceedings of the IEEE/CVF
  conference on computer vision and pattern recognition workshops}}.
  \bibinfo{pages}{0--0}.
\newblock


\bibitem[Maggiolino et~al\mbox{.}(2023)]%
        {maggiolino2023deep}
\bibfield{author}{\bibinfo{person}{Gerard Maggiolino}, \bibinfo{person}{Adnan
  Ahmad}, \bibinfo{person}{Jinkun Cao}, {and} \bibinfo{person}{Kris Kitani}.}
  \bibinfo{year}{2023}\natexlab{}.
\newblock \showarticletitle{Deep oc-sort: Multi-pedestrian tracking by adaptive
  re-identification}.
\newblock \bibinfo{journal}{\emph{arXiv preprint arXiv:2302.11813}}
  (\bibinfo{year}{2023}).
\newblock


\bibitem[Maglo et~al\mbox{.}(2022)]%
        {maglo2022efficient}
\bibfield{author}{\bibinfo{person}{Adrien Maglo}, \bibinfo{person}{Astrid
  Orcesi}, {and} \bibinfo{person}{Quoc-Cuong Pham}.}
  \bibinfo{year}{2022}\natexlab{}.
\newblock \showarticletitle{Efficient tracking of team sport players with few
  game-specific annotations}. In \bibinfo{booktitle}{\emph{Proceedings of the
  IEEE/CVF Conference on Computer Vision and Pattern Recognition}}.
  \bibinfo{pages}{3461--3471}.
\newblock


\bibitem[Mkhallati et~al\mbox{.}(2023)]%
        {Mkhallati2023SoccerNetCaptionDV}
\bibfield{author}{\bibinfo{person}{Hassan Mkhallati}, \bibinfo{person}{Anthony
  Cioppa}, \bibinfo{person}{Silvio Giancola}, \bibinfo{person}{Bernard Ghanem},
  {and} \bibinfo{person}{Marc~Van Droogenbroeck}.}
  \bibinfo{year}{2023}\natexlab{}.
\newblock \showarticletitle{SoccerNet-Caption: Dense Video Captioning for
  Soccer Broadcasts Commentaries}.
\newblock \bibinfo{journal}{\emph{2023 IEEE/CVF Conference on Computer Vision
  and Pattern Recognition Workshops (CVPRW)}} (\bibinfo{year}{2023}),
  \bibinfo{pages}{5074--5085}.
\newblock
\urldef\tempurl%
\url{https://api.semanticscholar.org/CorpusID:258049025}
\showURL{%
\tempurl}


\bibitem[Najafzadeh et~al\mbox{.}(2015)]%
        {najafzadeh2015multiple}
\bibfield{author}{\bibinfo{person}{Nima Najafzadeh}, \bibinfo{person}{Mehran
  Fotouhi}, {and} \bibinfo{person}{Shohreh Kasaei}.}
  \bibinfo{year}{2015}\natexlab{}.
\newblock \showarticletitle{Multiple soccer players tracking}. In
  \bibinfo{booktitle}{\emph{2015 The international symposium on artificial
  intelligence and signal processing (AISP)}}. IEEE, \bibinfo{pages}{310--315}.
\newblock


\bibitem[Scott et~al\mbox{.}(2022)]%
        {scott2022soccertrack}
\bibfield{author}{\bibinfo{person}{Atom Scott}, \bibinfo{person}{Ikuma Uchida},
  \bibinfo{person}{Masaki Onishi}, \bibinfo{person}{Yoshinari Kameda},
  \bibinfo{person}{Kazuhiro Fukui}, {and} \bibinfo{person}{Keisuke Fujii}.}
  \bibinfo{year}{2022}\natexlab{}.
\newblock \showarticletitle{SoccerTrack: A dataset and tracking algorithm for
  soccer with fish-eye and drone videos}. In
  \bibinfo{booktitle}{\emph{Proceedings of the IEEE/CVF Conference on Computer
  Vision and Pattern Recognition}}. \bibinfo{pages}{3569--3579}.
\newblock


\bibitem[Senocak et~al\mbox{.}(2018)]%
        {senocak2018part}
\bibfield{author}{\bibinfo{person}{Arda Senocak}, \bibinfo{person}{Tae-Hyun
  Oh}, \bibinfo{person}{Junsik Kim}, {and} \bibinfo{person}{In So~Kweon}.}
  \bibinfo{year}{2018}\natexlab{}.
\newblock \showarticletitle{Part-based player identification using deep
  convolutional representation and multi-scale pooling}. In
  \bibinfo{booktitle}{\emph{Proceedings of the IEEE Conference on Computer
  Vision and Pattern Recognition Workshops}}. \bibinfo{pages}{1732--1739}.
\newblock


\bibitem[Somers et~al\mbox{.}(2023)]%
        {somers2023body}
\bibfield{author}{\bibinfo{person}{Vladimir Somers},
  \bibinfo{person}{Christophe De~Vleeschouwer}, {and}
  \bibinfo{person}{Alexandre Alahi}.} \bibinfo{year}{2023}\natexlab{}.
\newblock \showarticletitle{Body part-based representation learning for
  occluded person Re-Identification}. In \bibinfo{booktitle}{\emph{Proceedings
  of the IEEE/CVF Winter Conference on Applications of Computer Vision}}.
  \bibinfo{pages}{1613--1623}.
\newblock


\bibitem[Suh et~al\mbox{.}(2018)]%
        {suh2018part}
\bibfield{author}{\bibinfo{person}{Yumin Suh}, \bibinfo{person}{Jingdong Wang},
  \bibinfo{person}{Siyu Tang}, \bibinfo{person}{Tao Mei}, {and}
  \bibinfo{person}{Kyoung~Mu Lee}.} \bibinfo{year}{2018}\natexlab{}.
\newblock \showarticletitle{Part-aligned bilinear representations for person
  re-identification}. In \bibinfo{booktitle}{\emph{Proceedings of the European
  conference on computer vision (ECCV)}}. \bibinfo{pages}{402--419}.
\newblock


\bibitem[Sun et~al\mbox{.}(2019)]%
        {sun2019deep}
\bibfield{author}{\bibinfo{person}{Ke Sun}, \bibinfo{person}{Bin Xiao},
  \bibinfo{person}{Dong Liu}, {and} \bibinfo{person}{Jingdong Wang}.}
  \bibinfo{year}{2019}\natexlab{}.
\newblock \showarticletitle{Deep high-resolution representation learning for
  human pose estimation}. In \bibinfo{booktitle}{\emph{Proceedings of the
  IEEE/CVF conference on computer vision and pattern recognition}}.
  \bibinfo{pages}{5693--5703}.
\newblock


\bibitem[Sun et~al\mbox{.}(2017)]%
        {PCB}
\bibfield{author}{\bibinfo{person}{Yifan Sun}, \bibinfo{person}{Liang Zheng},
  \bibinfo{person}{Yi Yang}, \bibinfo{person}{Qi Tian}, {and}
  \bibinfo{person}{Shengjin Wang}.} \bibinfo{year}{2017}\natexlab{}.
\newblock \showarticletitle{{Beyond Part Models: Person Retrieval with Refined
  Part Pooling (and a Strong Convolutional Baseline)}}.
\newblock \bibinfo{journal}{\emph{Lecture Notes in Computer Science (including
  subseries Lecture Notes in Artificial Intelligence and Lecture Notes in
  Bioinformatics)}}  \bibinfo{volume}{11208 LNCS} (\bibinfo{date}{nov}
  \bibinfo{year}{2017}), \bibinfo{pages}{501--518}.
\newblock
\showeprint[arxiv]{1711.09349}
\urldef\tempurl%
\url{http://arxiv.org/abs/1711.09349}
\showURL{%
\tempurl}


\bibitem[Sun et~al\mbox{.}(2018)]%
        {sun2018beyond}
\bibfield{author}{\bibinfo{person}{Yifan Sun}, \bibinfo{person}{Liang Zheng},
  \bibinfo{person}{Yi Yang}, \bibinfo{person}{Qi Tian}, {and}
  \bibinfo{person}{Shengjin Wang}.} \bibinfo{year}{2018}\natexlab{}.
\newblock \showarticletitle{Beyond part models: Person retrieval with refined
  part pooling (and a strong convolutional baseline)}. In
  \bibinfo{booktitle}{\emph{Proceedings of the European conference on computer
  vision (ECCV)}}. \bibinfo{pages}{480--496}.
\newblock


\bibitem[Tavassolipour et~al\mbox{.}(2013)]%
        {tavassolipour2013event}
\bibfield{author}{\bibinfo{person}{Mostafa Tavassolipour},
  \bibinfo{person}{Mahmood Karimian}, {and} \bibinfo{person}{Shohreh Kasaei}.}
  \bibinfo{year}{2013}\natexlab{}.
\newblock \showarticletitle{Event detection and summarization in soccer videos
  using bayesian network and copula}.
\newblock \bibinfo{journal}{\emph{IEEE Transactions on circuits and systems for
  video technology}} \bibinfo{volume}{24}, \bibinfo{number}{2}
  (\bibinfo{year}{2013}), \bibinfo{pages}{291--304}.
\newblock


\bibitem[Theagarajan and Bhanu(2020)]%
        {theagarajan2020automated}
\bibfield{author}{\bibinfo{person}{Rajkumar Theagarajan} {and}
  \bibinfo{person}{Bir Bhanu}.} \bibinfo{year}{2020}\natexlab{}.
\newblock \showarticletitle{An automated system for generating tactical
  performance statistics for individual soccer players from videos}.
\newblock \bibinfo{journal}{\emph{IEEE Transactions on Circuits and Systems for
  Video Technology}} \bibinfo{volume}{31}, \bibinfo{number}{2}
  (\bibinfo{year}{2020}), \bibinfo{pages}{632--646}.
\newblock


\bibitem[Thomas et~al\mbox{.}(2017)]%
        {thomas2017computer}
\bibfield{author}{\bibinfo{person}{Graham Thomas}, \bibinfo{person}{Rikke
  Gade}, \bibinfo{person}{Thomas~B Moeslund}, \bibinfo{person}{Peter Carr},
  {and} \bibinfo{person}{Adrian Hilton}.} \bibinfo{year}{2017}\natexlab{}.
\newblock \showarticletitle{Computer vision for sports: Current applications
  and research topics}.
\newblock \bibinfo{journal}{\emph{Computer Vision and Image Understanding}}
  \bibinfo{volume}{159} (\bibinfo{year}{2017}), \bibinfo{pages}{3--18}.
\newblock


\bibitem[Tong et~al\mbox{.}(2011)]%
        {tong2011automatic}
\bibfield{author}{\bibinfo{person}{Xiaofeng Tong}, \bibinfo{person}{Jia Liu},
  \bibinfo{person}{Tao Wang}, {and} \bibinfo{person}{Yimin Zhang}.}
  \bibinfo{year}{2011}\natexlab{}.
\newblock \showarticletitle{Automatic player labeling, tracking and field
  registration and trajectory mapping in broadcast soccer video}.
\newblock \bibinfo{journal}{\emph{ACM Transactions on Intelligent Systems and
  Technology (TIST)}} \bibinfo{volume}{2}, \bibinfo{number}{2}
  (\bibinfo{year}{2011}), \bibinfo{pages}{1--32}.
\newblock


\bibitem[Uchida et~al\mbox{.}(2021)]%
        {uchida2021automated}
\bibfield{author}{\bibinfo{person}{Ikuma Uchida}, \bibinfo{person}{Atom Scott},
  \bibinfo{person}{Hidehiko Shishido}, {and} \bibinfo{person}{Yoshinari
  Kameda}.} \bibinfo{year}{2021}\natexlab{}.
\newblock \showarticletitle{Automated Offside Detection by Spatio-Temporal
  Analysis of Football Videos}. In \bibinfo{booktitle}{\emph{Proceedings of the
  4th International Workshop on Multimedia Content Analysis in Sports}}.
  \bibinfo{pages}{17--24}.
\newblock


\bibitem[Vandeghen et~al\mbox{.}(2022)]%
        {Vandeghen2022SemiSupervisedTT}
\bibfield{author}{\bibinfo{person}{Renaud Vandeghen}, \bibinfo{person}{Anthony
  Cioppa}, {and} \bibinfo{person}{Marc~Van Droogenbroeck}.}
  \bibinfo{year}{2022}\natexlab{}.
\newblock \showarticletitle{Semi-Supervised Training to Improve Player and Ball
  Detection in Soccer}.
\newblock \bibinfo{journal}{\emph{2022 IEEE/CVF Conference on Computer Vision
  and Pattern Recognition Workshops (CVPRW)}} (\bibinfo{year}{2022}),
  \bibinfo{pages}{3480--3489}.
\newblock
\urldef\tempurl%
\url{https://api.semanticscholar.org/CorpusID:248178107}
\showURL{%
\tempurl}


\bibitem[Vats et~al\mbox{.}(2021)]%
        {vats2021multi}
\bibfield{author}{\bibinfo{person}{Kanav Vats}, \bibinfo{person}{Mehrnaz Fani},
  \bibinfo{person}{David~A Clausi}, {and} \bibinfo{person}{John Zelek}.}
  \bibinfo{year}{2021}\natexlab{}.
\newblock \showarticletitle{Multi-task learning for jersey number recognition
  in ice hockey}. In \bibinfo{booktitle}{\emph{Proceedings of the 4th
  International Workshop on Multimedia Content Analysis in Sports}}.
  \bibinfo{pages}{11--15}.
\newblock


\bibitem[Vats et~al\mbox{.}(2022)]%
        {vats2022ice}
\bibfield{author}{\bibinfo{person}{Kanav Vats}, \bibinfo{person}{William
  McNally}, \bibinfo{person}{Pascale Walters}, \bibinfo{person}{David~A
  Clausi}, {and} \bibinfo{person}{John~S Zelek}.}
  \bibinfo{year}{2022}\natexlab{}.
\newblock \showarticletitle{Ice hockey player identification via transformers
  and weakly supervised learning}. In \bibinfo{booktitle}{\emph{Proceedings of
  the IEEE/CVF Conference on Computer Vision and Pattern Recognition}}.
  \bibinfo{pages}{3451--3460}.
\newblock


\bibitem[Vats et~al\mbox{.}(2023)]%
        {vats2023player}
\bibfield{author}{\bibinfo{person}{Kanav Vats}, \bibinfo{person}{Pascale
  Walters}, \bibinfo{person}{Mehrnaz Fani}, \bibinfo{person}{David~A Clausi},
  {and} \bibinfo{person}{John~S Zelek}.} \bibinfo{year}{2023}\natexlab{}.
\newblock \showarticletitle{Player tracking and identification in ice hockey}.
\newblock \bibinfo{journal}{\emph{Expert Systems with Applications}}
  \bibinfo{volume}{213} (\bibinfo{year}{2023}), \bibinfo{pages}{119250}.
\newblock


\bibitem[Wang et~al\mbox{.}(2021)]%
        {wang2021pose}
\bibfield{author}{\bibinfo{person}{HongXia Wang}, \bibinfo{person}{Xiang Chen},
  {and} \bibinfo{person}{Chun Liu}.} \bibinfo{year}{2021}\natexlab{}.
\newblock \showarticletitle{Pose-guided part matching network via shrinking and
  reweighting for occluded person re-identification}.
\newblock \bibinfo{journal}{\emph{Image and Vision Computing}}
  \bibinfo{volume}{111} (\bibinfo{year}{2021}), \bibinfo{pages}{104186}.
\newblock


\bibitem[Wojke et~al\mbox{.}(2017)]%
        {wojke2017simple}
\bibfield{author}{\bibinfo{person}{Nicolai Wojke}, \bibinfo{person}{Alex
  Bewley}, {and} \bibinfo{person}{Dietrich Paulus}.}
  \bibinfo{year}{2017}\natexlab{}.
\newblock \showarticletitle{Simple online and realtime tracking with a deep
  association metric}. In \bibinfo{booktitle}{\emph{2017 IEEE international
  conference on image processing (ICIP)}}. IEEE, \bibinfo{pages}{3645--3649}.
\newblock


\bibitem[Xu et~al\mbox{.}(2021)]%
        {xu2021dual}
\bibfield{author}{\bibinfo{person}{Yunjie Xu}, \bibinfo{person}{Liaoying Zhao},
  {and} \bibinfo{person}{Feiwei Qin}.} \bibinfo{year}{2021}\natexlab{}.
\newblock \showarticletitle{Dual attention-based method for occluded person
  re-identification}.
\newblock \bibinfo{journal}{\emph{Knowledge-Based Systems}}
  \bibinfo{volume}{212} (\bibinfo{year}{2021}), \bibinfo{pages}{106554}.
\newblock


\bibitem[Yang et~al\mbox{.}(2023)]%
        {yang2023hard}
\bibfield{author}{\bibinfo{person}{Fan Yang}, \bibinfo{person}{Shigeyuki
  Odashima}, \bibinfo{person}{Shoichi Masui}, {and} \bibinfo{person}{Shan
  Jiang}.} \bibinfo{year}{2023}\natexlab{}.
\newblock \showarticletitle{Hard to track objects with irregular motions and
  similar appearances? make it easier by buffering the matching space}. In
  \bibinfo{booktitle}{\emph{Proceedings of the IEEE/CVF Winter Conference on
  Applications of Computer Vision}}. \bibinfo{pages}{4799--4808}.
\newblock


\bibitem[Yang et~al\mbox{.}(2021)]%
        {yang2021multi}
\bibfield{author}{\bibinfo{person}{Yukun Yang}, \bibinfo{person}{Ruiheng
  Zhang}, \bibinfo{person}{Wanneng Wu}, \bibinfo{person}{Yu Peng}, {and}
  \bibinfo{person}{Min Xu}.} \bibinfo{year}{2021}\natexlab{}.
\newblock \showarticletitle{Multi-camera sports players 3d localization with
  identification reasoning}. In \bibinfo{booktitle}{\emph{2020 25th
  International Conference on Pattern Recognition (ICPR)}}. IEEE,
  \bibinfo{pages}{4497--4504}.
\newblock


\bibitem[Zandycke et~al\mbox{.}(2022)]%
        {Zandycke2022DeepSportradarv1CV}
\bibfield{author}{\bibinfo{person}{Gabriel~Van Zandycke},
  \bibinfo{person}{Vladimir Somers}, \bibinfo{person}{Maxime Istasse},
  \bibinfo{person}{Carlo~Del Don}, {and} \bibinfo{person}{Davide Zambrano}.}
  \bibinfo{year}{2022}\natexlab{}.
\newblock \showarticletitle{DeepSportradar-v1: Computer Vision Dataset for
  Sports Understanding with High Quality Annotations}.
\newblock \bibinfo{journal}{\emph{Proceedings of the 5th International ACM
  Workshop on Multimedia Content Analysis in Sports}} (\bibinfo{year}{2022}).
\newblock
\urldef\tempurl%
\url{https://api.semanticscholar.org/CorpusID:251623078}
\showURL{%
\tempurl}


\bibitem[Zhang et~al\mbox{.}(2020)]%
        {zhang2020multi}
\bibfield{author}{\bibinfo{person}{Ruiheng Zhang}, \bibinfo{person}{Lingxiang
  Wu}, \bibinfo{person}{Yukun Yang}, \bibinfo{person}{Wanneng Wu},
  \bibinfo{person}{Yueqiang Chen}, {and} \bibinfo{person}{Min Xu}.}
  \bibinfo{year}{2020}\natexlab{}.
\newblock \showarticletitle{Multi-camera multi-player tracking with deep player
  identification in sports video}.
\newblock \bibinfo{journal}{\emph{Pattern Recognition}}  \bibinfo{volume}{102}
  (\bibinfo{year}{2020}), \bibinfo{pages}{107260}.
\newblock


\bibitem[Zhang et~al\mbox{.}(2022)]%
        {zhang2022bytetrack}
\bibfield{author}{\bibinfo{person}{Yifu Zhang}, \bibinfo{person}{Peize Sun},
  \bibinfo{person}{Yi Jiang}, \bibinfo{person}{Dongdong Yu},
  \bibinfo{person}{Fucheng Weng}, \bibinfo{person}{Zehuan Yuan},
  \bibinfo{person}{Ping Luo}, \bibinfo{person}{Wenyu Liu}, {and}
  \bibinfo{person}{Xinggang Wang}.} \bibinfo{year}{2022}\natexlab{}.
\newblock \showarticletitle{Bytetrack: Multi-object tracking by associating
  every detection box}. In \bibinfo{booktitle}{\emph{European Conference on
  Computer Vision}}. Springer, \bibinfo{pages}{1--21}.
\newblock


\bibitem[Zhang et~al\mbox{.}(2021)]%
        {zhang2021fairmot}
\bibfield{author}{\bibinfo{person}{Yifu Zhang}, \bibinfo{person}{Chunyu Wang},
  \bibinfo{person}{Xinggang Wang}, \bibinfo{person}{Wenjun Zeng}, {and}
  \bibinfo{person}{Wenyu Liu}.} \bibinfo{year}{2021}\natexlab{}.
\newblock \showarticletitle{Fairmot: On the fairness of detection and
  re-identification in multiple object tracking}.
\newblock \bibinfo{journal}{\emph{International Journal of Computer Vision}}
  \bibinfo{volume}{129} (\bibinfo{year}{2021}), \bibinfo{pages}{3069--3087}.
\newblock


\bibitem[Zheng et~al\mbox{.}(2015)]%
        {zheng2015scalable}
\bibfield{author}{\bibinfo{person}{Liang Zheng}, \bibinfo{person}{Liyue Shen},
  \bibinfo{person}{Lu Tian}, \bibinfo{person}{Shengjin Wang},
  \bibinfo{person}{Jingdong Wang}, {and} \bibinfo{person}{Qi Tian}.}
  \bibinfo{year}{2015}\natexlab{}.
\newblock \showarticletitle{Scalable person re-identification: A benchmark}. In
  \bibinfo{booktitle}{\emph{Proceedings of the IEEE international conference on
  computer vision}}. \bibinfo{pages}{1116--1124}.
\newblock


\end{thebibliography}

\end{document}